\documentclass[]{new_tlp}

\usepackage{microtype}
\usepackage[utf8]{inputenc}
\usepackage{amsmath,amssymb}
\usepackage{graphicx}
\usepackage{color}
\usepackage{url}
\usepackage{multirow}
\usepackage{listings}
\lstdefinelanguage{clingo}{
  keywordstyle=[1]\usefont{OT1}{cmtt}{m}{n},%
  keywordstyle=[2]\textbf,%
  keywordstyle=[3]\usefont{OT1}{cmtt}{m}{n},
  alsoletter={\#,\&},%
  keywords=[1]{not,from,import,def,if,else,return,while,break,and,or,for,in,del,and,class},%
  keywords=[2]{\#const,\#show,\#minimize,\#base,\#theory,\#count,\#external,\#program,\#script,\#end,\#heuristic,\#edge,\#project,\#show},%
  keywords=[3]{&,&dom,&sum,&diff,&show,&minimize},%
  morecomment=[l]{\#\ },%
  morecomment=[l]{\%\ },%
  commentstyle={\color{darkgray}}%
}

\newcommand{\Adom}{\textbf{A}}
\newcommand{\Bdom}{\textbf{B}}
\newcommand{\Cdom}{\textbf{C}}
\newcommand{\Mdom}{\textbf{M}}
\newcommand{\AMdom}{\ensuremath{\mathbf{A^{\!M}}}}
\newcommand{\BMdom}{\ensuremath{\mathbf{B^M}}}
\newcommand{\CMdom}{\ensuremath{\mathbf{C^M}}}
\newcommand{\MdomAsg}{\ensuremath{\mathbf{M_{a}}}}
\newcommand{\AMdomAsg}{\ensuremath{\mathbf{A^{\!M}_{a}}}}
\newcommand{\BMdomAsg}{\ensuremath{\mathbf{B^M_{a}}}}
\newcommand{\CMdomAsg}{\ensuremath{\mathbf{C^M_{a}}}}
\newcommand{\HN}[1]{{\fontseries{b}\selectfont #1}}

\newcommand{\clingosplit}{\sysfont{clingo}\ensuremath{_{xy}}}

\newcommand{\sysfont}{\textit}

\newcommand{\asprilo}{\sysfont{asprilo}}

\newcommand{\clingcon}{\sysfont{clingcon}}
\newcommand{\clingo}{\sysfont{clingo}}
\newcommand{\clingod}[1]{\clingo\textnormal{[}\textsc{#1}\textnormal{]}}

\newcommand{\python}{Python}

\title[Experimenting with robotic intra-logistics domains]{Experimenting with robotic intra-logistics domains}

\author[Gebser et al.]{%
Martin Gebser,
Philipp Obermeier,
Thomas Otto,
Torsten Schaub\\
University of Potsdam, Germany
\and
Orkunt Sabuncu\\
TED University, Ankara, Turkey
\and
Van Nguyen,
Tran Cao Son \\
New Mexico State University, Las Cruces, USA
}

\submitted{[n/a]}
\revised{[n/a]}
\accepted{[n/a]}

\begin{document}

\maketitle

\begin{abstract}
We introduce the \asprilo\footnote{\asprilo{} stands for \emph{Answer Set Programming for robotic intra-logistics}.}
framework
to facilitate experimental studies of approaches addressing complex dynamic applications.
For this purpose, we have chosen the domain of robotic intra-logistics.
This domain is not only highly relevant in the context of today's fourth industrial revolution
but it moreover combines a multitude of challenging issues within a single uniform framework.
This includes
multi-agent planning,
reasoning about action, change,
resources,
strategies,
etc.
In return, \asprilo{} allows users to study alternative solutions as regards effectiveness and scalability.
Although \asprilo{} relies on Answer Set Programming and \python, it is readily usable by any system
complying with its fact-oriented interface format.
This makes it attractive for benchmarking and teaching well beyond logic programming.
More precisely,
\asprilo{} consists of a versatile benchmark generator, solution checker and visualizer as well as
a bunch of reference encodings featuring various ASP techniques.
Importantly,
the visualizer's animation capabilities are indispensable for complex scenarios like intra-logistics in order to inspect valid as well as invalid solution candidates.
Also, it allows for graphically editing benchmark layouts that can be used as a basis for generating benchmark suites.

\medskip\noindent
{\em Under consideration for publication in Theory and Practice of Logic Programming (TPLP)}
\end{abstract}

\section{Introduction}
\label{sec:introduction}

Answer Set Programming (ASP;~\citeNP{lifschitz99b}) has come a long way,
starting as a semantics for logic programming,
over having increasingly performant systems,
to a growing number of significant applications
in academia and industry.
However,
this development is threatened by a lack of complex benchmark scenarios
mimicking the needs of real-world applications.
In contrast to many available benchmark suites,
often supplied by automatic instance generators,
real-world applications are rarely disseminated,
either because they are classified
or come only with a handful of instances.
Another commonality of existing benchmark suites is that they are kept simple,
stick to basic ASP, and usually feature at most a single specifics,
so that they can be processed by as many systems as possible.
However, this is in contrast to many real-world applications whose solution requires the integration
of multiple types of knowledge and forms of reasoning.
Last but not least,
a feature distinguishing ASP from all other solving paradigms is its versatility,
which is best put in perspective by solving multi-faceted problems.

We fear that
the lack of complex benchmark scenarios becomes a major bottleneck in ASP's progression towards real-world applications,
and hence that more and more should be made available to our community.
As a first step to overcome this problem,
we have identified the domain of robotic intra-logistics,
a key domain in the context of the fourth industrial revolution,
as witnessed by Amazon's Kiva, GreyOrange's Butler, and Swisslog's CarryPick systems.%
\footnote{\url{www.amazonrobotics.com}, \url{www.greyorange.com/products/butler}, \url{www.swisslog.com/carrypick}}
All of them aim at automatizing warehouse operations by using robot vehicles
that drive underneath mobile shelves and deliver them to picking stations.
From there, workers pick and place the requested items in shipping boxes.
Apart from the great significance of this real-world domain,
our choice is motivated by several aspects.
First of all,
the warehouse layout is grid-based and thus provides a suitable abstraction for modeling robot movements in ASP.
Moreover,
the domain offers a great variety of manifold problem scenarios that can be put together in an increasingly complex way.
For instance,
one may start with single or multi-robot path-finding scenarios induced by a set of orders
that are accomplished by using robots for transporting shelves to picking stations.
This can be extended in various ways, for example, by
adding shelf handling and delivery actions,
considering order lines with multiple product items,
keeping track of the number of ordered and/or stored product items,
modeling energy consumption and charging actions,
taking into account order frequencies, supplies, and priorities,
striving for effective layouts featuring dedicated locations,
like highways or storage areas,
and so on.
Finally, the domain is extremely well-suited for producing scalable benchmarks 
by varying layouts, robots, shelves, orders, product items, etc.
Inspired by this,
we have developed the benchmark environment \asprilo{} consisting of four parts
(i)   a benchmark generator,
(ii)  a solution checker,
(iii) a benchmark and solution visualizer, and
(iv)  a variety of reference encodings.
The design of \asprilo{} was driven by the desire to create an
easily configurable and extensible framework that allows for
generating scalable, standardized benchmark suites that
can be visualized with and without a corresponding solution.
Correctness can be established via a modular solution checker.
The accompanying reference encodings may serve as exemplary bases for extended encodings addressing more complex scenarios.
We use them in Section~\ref{sec:experiments} to illustrate \asprilo's usage and value for benchmarking.
In particular,
we show how \asprilo{} may allow us to understand the virtues of different ASP techniques.
The \asprilo{} framework is freely available
at~\url{https://potassco.org/asprilo}.


\section{\asprilo's benchmark generation component}
\label{sec:generation}

\subsection{Problem domains}
\label{sec:problems}
All \asprilo{} domains aim at capturing abstractions of intra-logistics systems
involving multiple robots, shelves, and stations,
placed in a warehouse environment,
along with a set of orders.
We begin with a description of the common setting.

Given a set of orders,
the idea is to have robots bring shelves containing requested products to picking stations
until all orders are satisfied.
The \emph{warehouse} floor is laid out as a two-dimensional grid of squares.
A square can be declared as \emph{highway}, \emph{picking station},%
\footnote{Other dedicated stations, like charging or supplying stations, are foreseen in future editions of \asprilo.}
or \emph{storage location}.
Only the latter can be occupied by shelves without being carried by robots.
A \emph{shelf} stores products in certain quantities,
and a \emph{picking station} accepts product units requested by orders.
Both occupy a single grid square.
A square may be occupied by at most one robot and one shelf, no matter whether the robot carries the shelf or not.
Hence, robots may pass or park underneath shelves, unless they carry one.
\emph{Highways} are used to declare transit areas for shelf carrying robots;
they are neither shelf storage locations nor parking lots for idle robots.%
\footnote{We are currently not distinguishing parking areas for robots and use storage locations for simplicity.}

Per time step,
a \emph{robot} can perform at most one action,
either\footnote{More actions, like battery charging or restocking, are planned but not yet part of \asprilo.}
\emph{move} to an adjacent square,
\emph{pick up} or
\emph{put down} a shelf, or
\emph{deliver} (products on) a shelf to a picking station.
Robots must not swap squares at a time step.
A robot may pick up a shelf, provided that both are on the same square and it does not yet carry a shelf;
it may put down a shelf, if it carries the shelf.
Thus,
a robot can pick up a shelf, carry it while moving, and put it down eventually.
However,
a robot may never put down a shelf outside the storage area.
A robot carrying a shelf may deliver product units from the shelf at a picking station,%
\footnote{In reality, this is usually a passive action since a person or a robot picks items from the shelf.}
provided it
occupies the corresponding square and
the product units are part of an order line processed at this station.
The specific arrangement of deliveries distinguishes several \asprilo{} domains and is made precise below.

An \emph{order} is a non-empty set of \emph{order lines},
which are requests for products in certain quantities
to be delivered to a given picking station.
For simplicity,\footnote{Supply management is another possible future extension of \asprilo.}
all requested products are available in sufficient quantity on the shelves in the warehouse.
An order line is fulfilled if the requested amount of product units has been delivered to
the designated picking station.
An order is fulfilled if all its order lines are fulfilled.
In the current setting, all orders are provided initially (instead of arriving over time).
In these terms,
the overall \emph{goal} can be rephrased to find a parallel plan of robot actions that fulfills all orders.

In what follows,
we introduce selected problem domains of \asprilo.
We begin with the (currently) most general setting and gradually refine it afterwards to obtain simplified variants.
In the general \asprilo{} domain \Adom,
deliveries fully account for the quantities of product units taken off a shelf.
A delivery action deals with a single order line;
the number of transferred units can neither exceed
the quantity available on the shelf nor
the requested amount of units.
Thus, multiple order lines require multiple deliveries,
and one order line may require units from several shelves.
%

The \asprilo{} domain \Bdom{} simplifies \Adom{} by ignoring product quantities.
A delivery action still deals with a single order line
but only the requested product must be on the shelf (and remains after delivery),
no matter how many units are available or requested.
The \asprilo{} domain \Cdom{} further simplifies \Adom{} and \Bdom{}
by making one delivery action deal at once with all pending order lines matching products on the shelf.
Thus, in this setting, one deliver action may fulfill several orders at a picking station all at once.

The \asprilo{} domain \Mdom{} is a drastic simplification
designed to resemble multi-agent path finding scenarios known from the literature
(cf.\ Section~\ref{sec:related}).
This domain features only \emph{move} actions and the only concern is to find a parallel plan that positions each robot under a specific shelf.
Hence, shelves remain on the spot and no products are delivered.
To accommodate this scenario in the \asprilo{} framework,
we begin by equipping each shelf with a unique product.
The number of robots and orders is aligned,
and each order lines up a unique product.
All ordered products are stocked, at most one per shelf.
An order (line) is fulfilled if some robot is located under the shelf holding a requested product.
Note that robots are not pre-assigned to shelves in \Mdom, as common in certain multi-agent path finding scenarios.
The addition of such an assignment constitutes yet another variant,
and corresponds to the traditional MAPF problem, as detailed in Section~\ref{sec:related}.

To bridge the gap between scenarios \Adom, \Bdom, \Cdom{} and \Mdom,
we also consider the restriction of the former to the specific shelf and order alignment used in the latter.
The resulting scenarios \AMdom, \BMdom, and \CMdom{} thus deal with full-fledged delivery scenarios
but only singleton orders and shelves with unique product units.
This allows us to experiment with the general functionality of \Adom, \Bdom, and \Cdom{} in the rather restricted warehouse setting of \Mdom.

Clearly further constraints can be imposed, leading to a wide variety of \asprilo{} scenarios.

\subsection{Instance format}
\label{sec:instance}
Our instance format follows the well-known object-attribute-value scheme
that provides a generic and extensible knowledge representation format.
More precisely, we represent each initial item as a fact of binary predicate \lstinline{init/2} of form
\begin{lstlisting}[numbers=none,mathescape,,belowskip=2pt,aboveskip=2pt,basicstyle=\ttfamily]
   init(object($T$,$I$),value($A$,$V$)).
\end{lstlisting}
where
$T$ is an object type,
$I$    a (relative) object identifier,
$A$    an attribute, and
$V$    its value.
The current object types along with their admissible attributes are as follows:
{\topsep1pt%
\begin{center}
  \begin{tabular}{l@{\qquad}l}
    \texttt{node}           & \texttt{at/2}\\
    \texttt{highway}        & \texttt{at/2}\\
    \texttt{robot}          & \texttt{at/2}, \texttt{carries/1}\\
    \texttt{shelf}          & \texttt{at/2}\\
    \texttt{pickingStation} & \texttt{at/2}\\
    \texttt{product}        & \texttt{on/2}\\
    \texttt{order}          & \texttt{line/2}, \texttt{pickingStation/1}
  \end{tabular}
\end{center}}
The trailing number indicates the number of arguments of each attribute.
For instance,
the fact
`\lstinline{init(object(robot,34),value(at,(2,3))).}'
represents that robot 34 is at position (2,3).
Note that all but the last type in the above list are part of the initial layout of a warehouse and its inventory.
Each \texttt{node} has a unique identifier as well as a two-dimensional position.
Optionally, a position can be declared as a \texttt{highway}.
The major entities, namely, robots, shelves, and picking stations, have (initial) positions.
Also, a \texttt{robot} may carry a \texttt{shelf}.
The type \texttt{product} gives the inventory by listing how many product items are stored on a \texttt{shelf}.
Finally,
each \texttt{order} consists of one or more order lines, each fixing how many product items are requested,
and its destination, namely, the picking station where the order is put together.

\subsection{Instance generator}
\label{sec:generator}
%
The instance generator of \asprilo{} allows for generating benchmark sets with variable features and degrees of difficulty.
Each instance is expressed in the format described above;
its name is standardized to reflect its major characteristics,
viz.\
the grid dimension and the
numbers of
nodes, robots, shelves, picking stations, products, their quantities, and orders.
To guarantee reproducibility the header of each benchmark file gives the generator's version and instruction used to generate it.

The generator currently supports three types of layouts.
A structured one, resembling real-world editions,
a randomized one, and
customized ones.
In the structured layout,
picking stations and robots are initially placed in the upper and lower row, respectively.
Shelves are placed in rectangular clusters reachable via surrounding highways.
An example is shown in Figure~\ref{fig:screenshot:asprilo}.
No such structure is foreseen in a randomized layout.
Finally,
layouts can be handcrafted with \asprilo's graphical editor
(and afterwards populated with the instance generator).
Moreover,
the instance generator warrants that picking stations are not placed on highways
and optionally that all admissible shelf positions are directly reachable from highway positions.

%
The instance generator is implemented via ASP.
However,
to handle the huge space of warehouse configurations and incoming orders,
its implementation relies on multi-shot ASP solving and is controlled by the \python{} API of \clingo~\cite{gekakasc17a}.

Although instances can be generated in a single solving step,
benchmark sets are much more effectively generated incrementally
by separating different sources of combinatorics.
To this end,
the generator relies upon an ASP encoding that is compartmentalized by means of \clingo's \lstinline{#program} directive
(cf.~\cite{gekakasc17a}).
Each such subprogram encapsulates a dimension of our scenario.
Exemplary dimensions are
the type of layout,
the placement of shelves on admissible positions, and
the one of products units on shelves (while respecting their global distribution).
At first,
the compartmentalization into subprograms allows us to configure the ultimate logic program according to the user's input.
A typical example is the selection among layout alternatives.
Moreover,
the incremental approach takes advantage of compartmentalization by successively building benchmark instances
in order to reduce combinatorics.
At first,
the grid is fixed, and robots and picking stations are placed.
Alternatively, a handcrafted layout may be used (see below).
Then, in turn,
shelves,
product units,
and finally
orders are generated.
The overall process is controlled via \clingo's \python{} API that successively grounds and solves the subprograms in focus.
The incremental approach reduces combinatorics in two interesting ways.
First, previous steps allow us to delineate the variability of objects in later steps.
For instance,
only admissible positions must be considered when placing shelves;
similarly products must only be distributed among placed shelves.
Second,
we use configurable threshold-based domain splitting.
To illustrate this, consider shelf placement and suppose we need to place 55 shelves on 100 positions.
A splitting threshold of 20 makes the generator first choose 20 among 100 possible shelf positions.
Then, the generator places 40 shelves on the 100 positions, but fixes the ones determined previously
(by adding the corresponding facts);
hence only 20 are effectively placed.
And finally 55 shelves are placed, among which 40 are already fixed.
Note that the selection of the respective threshold is a trade-off.
Using the full set of objects is often infeasible,
and although very small sets are easily treated,
they produce many repetitive grounding and solving calls.

Finally, we mention that the generator features several ways of producing varied benchmark instances for the same configuration.

Now
to accommodate the variety of its benchmark scenarios,
\asprilo's instance generator is highly configurable
through a plethora of 68~options 
for specifying output instances, most notably the grid size, layout style, and object quantities.
For example,
the instance depicted in Fig.~\ref{fig:screenshot:asprilo} is created with the command
\begin{lstlisting}[numbers=none,belowskip=2pt,aboveskip=2pt,basicstyle=\small\ttfamily,label={listing:invocation}]
  gen -x 19 -y 9 -X 5 -Y 2 -p 3 -s 45 -r 6 -P 180 -u 540 -o 12 -H
\end{lstlisting}
where parameter
\lstinline{-x} and \lstinline{-y} fix the dimensions of the grid,
\lstinline{-X} and \lstinline{-Y}     the dimensions of the storage zones,
\lstinline{-p} the number of picking stations,
\lstinline{-s} the number of shelves,
\lstinline{-r} the number of robots,
\lstinline{-P} the number of products,
\lstinline{-u} the number of product units,
\lstinline{-o} the number of orders, and
\lstinline{-H} invokes the structured layout common in industrial settings,
involving highways and rectangular storage zones (along with the aforementioned systematic placement of robots and picking stations).

To produce more than one instance at a time,
the desired number of instances can be given with \lstinline{-N}.
Further,
option \lstinline{-I} generates instances incrementally via multi-shot solving, as detailed above.
For example,
to incrementally generate 10 instances with our previous call,
we just append \lstinline{-N 10} and \lstinline{-I} to the command line above.

Besides single invocations,
\asprilo{} also offers batch processing to generate multiple instance sets based on a list of configurations.%
\footnote{The YAML vocabulary of batch files is described at \url{http://potassco.org/asprilo}.}
Moreover,
the batch file syntax not only allows us to process several configurations consecutively
but also combinations thereof.
That is,
we can first stipulate a common (partial) preset of configuration parameters and afterwards derive
more specialized sub-configurations from it by stating further parameter settings.
For instance,
we can initially pre-define the grid size, number of picking stations, shelves,
products, and orders for all instances.
Based on that preset, we can then define sub-configurations that state different amounts of robots.
With it,
the creation of complex instance sets can be fully defined (in a reproducible way)
by a single batch file without the need for auxiliary scripts or manual curation.

\subsection{Solution checker}
\label{sec:checker}
A solution to an \asprilo{} scenario is a parallel plan represented by facts of the form
\begin{lstlisting}[numbers=none,mathescape,belowskip=2pt,aboveskip=2pt,basicstyle=\ttfamily]
   occurs(object($T$,$I$),action($A$,$V$)).
\end{lstlisting}
where
$T$ is an object type,
$I$    an object identifier,
$A$    an action name, and
$V$    a tuple of terms capturing the action's arguments.
A \lstinline{move} action takes cardinal points represented by
\lstinline{(0,1)},
\lstinline{(1,0)},
\lstinline{(0,-1)}, and
\lstinline{(-1,0)}.
Actions \lstinline{pickup} and \lstinline{putdown} have no arguments, expressed by \lstinline{()}.
In general, \lstinline{deliver} takes a triple \lstinline[mathescape]{($O$,$A$,$N$)},
where
$O$ is an order,
$A$ the delivered product, and
$N$ its quantity.
The latter is ignored in scenarios \Bdom{} and \Cdom{}.

To verify the correctness of such plans,
\asprilo{} features a modular solution checker written in ASP.
Depending on the scenario,
the corresponding constraints are put together to form the appropriate checker.
%
Apart from being modular,
this offers an easily extensible approach for accommodating future extensions of \asprilo.

Moreover,
the checker's encoding offers diagnostic support indicating the respective violation.
For this,
it captures plan inconsistencies by returning atoms of the form
\lstinline[mathescape]{err($F$,$C$,$P$)}
where
$F$ gives the name of the file defining the error,
$C$ holds the name of the violated constraint, and
$P$ specifies the parameters involved in its violation.

For example, suppose the checker returns the atom
\begin{lstlisting}[numbers=none,belowskip=2pt,aboveskip=2pt,basicstyle=\ttfamily]
   err(goal,unfilledOrder,(3,3,1,11))
\end{lstlisting} 
This indicates that there is still an unfulfilled order at the end of the plan at step \lstinline{11},
namely, that order \lstinline{3} still requires \lstinline{1} unit of product \lstinline{3}.
Further details are obtained by inspecting the rule in file \lstinline{goal.lp},
which defines this specific instance of \lstinline{err/3}.


\section{\asprilo's visualization component}
\label{sec:visualizer}
%
\begin{figure}
  \centering
  \includegraphics[width=0.9\textwidth]{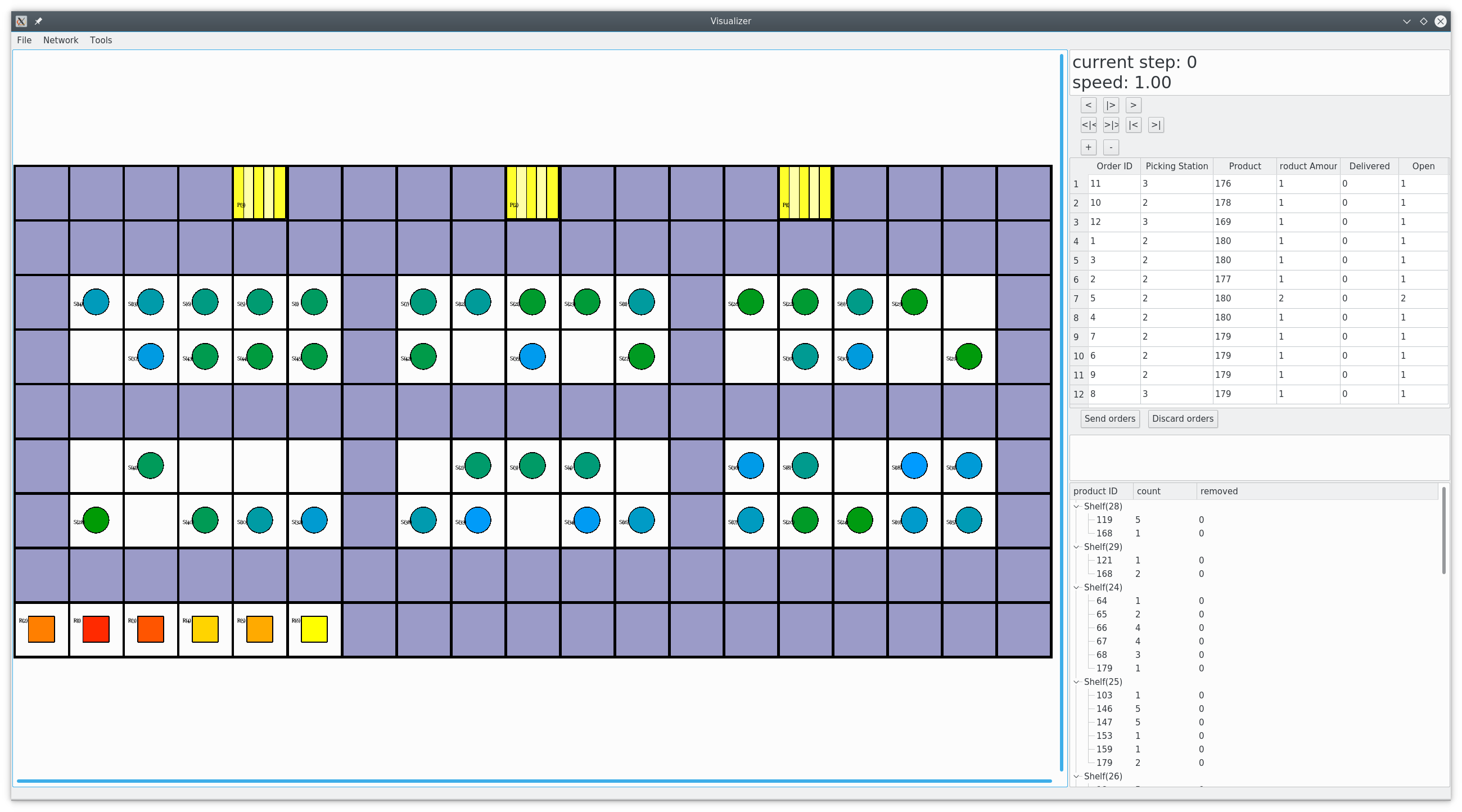}
  \caption{Exemplary \asprilo{} screenshot:
    The main window gives a warehouse layout; no plan is loaded.
    Picking stations are represented by striped yellow squares,
    shelves by solid circles, and
    robots by solid squares.
    Highways are brought out in purple.
    The side windows provide
    controls for plan animation,
    and give details about the current orders and
    the warehouse inventory.
  }
  \label{fig:screenshot:asprilo}
\end{figure}
%
The visualization component of \asprilo{} serves two major purposes,
the animation of solution candidates and
the editing of benchmark templates.
Such a template ranges from an empty scenario, over partial ones, to fully detailed warehouse environments.
The latter can be accompanied by a (possibly invalid) parallel plan that can be animated in the supplied environment.
Both are given in the previously described formats and can be passed to \asprilo's visualizer either through files given on the command line,
read from standard input, or loaded via a menu entry.

To begin with, let us sketch \asprilo's animation capabilities.
First of all, note that it is not coupled with the aforementioned solution checker.
This is important to allow for debugging invalid solution candidates.

The states of an \asprilo{} scenario are captured in a multi-modal way.
First, the main window captures the position of objects as shown in Fig.~\ref{fig:screenshot:asprilo}.
(In addition, robots carrying and being under shelves are indicated by distinct overlays of the respective icons.)
Second, a side window gives a table reflecting the current status of all orders
(reflecting closed and open order lines along with the delivered and missing product units).
Finally, another side window lists the warehouse's inventory by displaying the product units from the shelves.
The latter is also obtained when hovering with the mouse over the icon of a shelf in the main window.
Similarly, information is obtainable on robots and picking stations.

State transitions are animated to simulate smooth robot movements.
They can be traced via yet another side window holding the plan and highlighting all upcoming actions. 
The exploration of the plan at hand is done via common media controls (viz.\ play, pause, fast forward, etc.)
to enact and change the animation.
It comes without saying that the aforementioned state descriptions are updated accordingly.

Finally, we mention that scenarios and plans can be discharged and reloaded;
also their textual representation can be edited.

The second major feature of \asprilo's visualizer is its graphical scenario editor.
With it,
one may start from scratch with designing benchmark instances or templates,
or modify existing designs.
Templates can be further enriched by means of the above benchmark generator,
leading to entire benchmark series.
The functionality of \asprilo's graphical editor allows for
changing the grid dimensions,
adding and removing robots, shelves, and stations,
changing product units on shelves,
declaring squares as highway or storage,
and even entirely removing squares from the grid.
The latter is of particular interest when designing non-uniform layouts,
featuring obstacles such as corridors or pillars.
Apart from the change of dimensions,
all above functionalities are accomplished by mouse control.
Also, objects can be moved by drag and drop.
Similarly, orders can be edited textually in the corresponding side window.

A third feature, skipped for brevity, allows for connecting \asprilo's visualizer with solving components over the network.
%

\section{Exemplary encodings}
\label{sec:encoding}

We provide reference encodings for the various \asprilo{} scenarios at \url{https://potassco.org/asprilo}.
For brevity,
we illustrate here only our basic approach, viz.\ \Mdom, and merely sketch its extensions to more complex settings afterwards.
Common to all our encodings is a mapping from \asprilo's input format to a more comprehensible format.
This is addressed with rules like `\lstinline{robot(R) :- init(object(robot,R),_).}' gathered in \texttt{input.lp} (cf.~Line~1 in Listing~\ref{enc:M}).
Another commonality of our encodings concerns the separation of the action theory from its goal specification,
reflected by Listing~\ref{enc:M} and~\ref{enc:M:O}.

Listing~\ref{enc:M} gives a bounded encoding for generating plans in \asprilo{} scenario \Mdom.
\begin{lstlisting}[floatplacement=t,label=enc:M,basicstyle=\scriptsize\ttfamily,captionpos=b,caption={Encoding for \asprilo{} domain \Mdom.}]
#include "../input.lp".

time(1..horizon).

direction((X,Y)) :- X=-1..1, Y=-1..1, |X+Y|=1.
nextto((X,Y),(X',Y'),(X+X',Y+Y')) :- position((X,Y)), direction((X',Y')), position((X+X',Y+Y')).

{ move(R,D,T) : direction(D) } 1 :- isRobot(R), time(T).

position(R,C,T) :- move(R,D,T), position(R,C',T-1),     nextto(C',D,C).
                :- move(R,D,T), position(R,C ,T-1), not nextto(C ,D,_).

position(R,C,T) :- position(R,C,T-1), not move(R,_,T), isRobot(R), time(T).

moveto(C',C,T) :- nextto(C',D,C), position(R,C',T-1), move(R,D,T).
 :- moveto(C',C,T), moveto(C,C',T).

 :- { position(R,C,T) : isRobot(R) }  > 1, position(C), time(T).
\end{lstlisting}
The plan length is fixed by \lstinline{horizon} in Line~3.
Line~5 gives the four cardinal directions,
used in Line~6 to represent all transitions on the warehouse grid.
Predicate~\lstinline{position/1} provides all grid coordinates,
gathered in \texttt{input.lp} in analogy to the above.
Line~8 generates all possible robot moves per time step.
The corresponding effects and preconditions are expressed in Line~10 and~11, respectively.
The inertia of robot positions, captured by~\lstinline{position/3}, is given in Line~13.
The remaining rules express state constraints.
Lines~15 and~16 stipulate that robots must not swap positions,
and Line~18 forbids more than one robot per square.

The drastically simplified concept of order fulfillment in \Mdom{} is encoded in Listing~\ref{enc:M:O}.
\begin{lstlisting}[floatplacement=t,label=enc:M:O,basicstyle=\scriptsize\ttfamily,captionpos=b,caption={Encoding for order fulfillment in \asprilo{} domain \Mdom.}]
processed(A) :- ordered(O,A), shelved(S,A), isRobot(R), position(S,C,0), position(R,C,horizon).

 :- ordered(O,A), not processed(A).
\end{lstlisting}
An (ordered) product is \lstinline{processed} according to Line~1,
if some robot is parked at the last time step under a shelf holding the product.
And the goal is fulfilled, if all ordered products have been \lstinline{processed}.

We provide several alternative encodings for \asprilo{} scenario \Mdom,
which mainly differ in how positions are represented.
At first,
we propose an encoding featuring a separate representation of coordinates.
That is, rather than using predicate~\lstinline{position/3} whose second argument is a pair~\lstinline{(X,Y)},
we use two predicates~\lstinline{positionX/3} and~\lstinline{positionY/3} in which the second argument
contains only one coordinate, namely~\lstinline{X} or~\lstinline{Y}, respectively.
This representation is similar to the one used for Constrained ASP (with \clingcon~\cite{bakaossc16a}),
where both coordinates are represented functionally via integer variables~\lstinline{positionX/2} and~\lstinline{positionY/2}
whose values give the \lstinline{X} and~\lstinline{Y} coordinate, respectively.
The same conceptualization is used for the fourth variant relying upon ASP enhanced with difference constraints (with \clingod{dl}~\cite{jakaosscscwa17a})
Also, we provide incremental versions of all but the last variant (since it is not multi-shot capable).

An even greater variety of encodings is made available for scenarios \Adom, \Bdom, \Cdom.
Their common action theory deals with robot moves and pick up and put down actions.
However, in addition to alternative representations of robots and shelves,
we consider alternative representations of product units along with their combinations.
The difference between \Adom, \Bdom, and \Cdom{} is handled separately via different concepts of order fulfillment and thus different forms of
delivery actions.
Hence, as above, each setting boils down to two files, a specific encoding of the common action theory and another accounting for order fulfillment.


\section{Experimental evaluation}
\label{sec:experiments}

To showcase \asprilo's utility for experimenting with challenges emerging from robotic intra-logistics,
we conducted some initial case studies using \asprilo{} domains \Mdom{} and \CMdom, \BMdom, \AMdom{} (in increasing difficulty).
We give a detailed account of our experiments, including encodings, instance sets, and results at \url{http://potassco.org/asprilo}.

Our experimental analysis focuses on the following three questions:
\begin{enumerate}
\item
What is the impact of different representations of grid positions?
\item
What is the impact of increasingly complex domains?
\item
What is the impact of decoupling sources of combinatorics?
\end{enumerate}
We investigate the first question by comparing the four alternative encodings of positions discussed in Section~\ref{sec:encoding}.
This is carried out in \asprilo{} domain \Mdom, since positions only change by movement.
For the second question,
we study the addition of actions and more complex goal conditions,
and focus on domains \CMdom, \BMdom, \AMdom{} as direct extensions of \Mdom.
The outcome of this question also reflects on the difference between
traditionally studied, sparse multi-agent path finding (MAPF) scenarios (cf.~Section~\ref{sec:related})
and richer warehouse scenarios as provided by \asprilo.
For the third question,
we focus on separating task assignment and planning.
To this end,
we proceed in two steps.
First, robots are assigned to transport certain shelves and to deliver at certain picking stations.
This assignment is done subject to several optimization criteria that aim at selecting economic plans.
Then, the resulting assignment is added to the original action theory along with constraints enforcing
that actions are executed according to the given assignment.
%
The use of explicit task assignments is related to the difference between anonymous and non-anonymous MAPF (cf.~Section~\ref{sec:related}).

To minimize the degrees of freedom,
we consider instances of \Mdom{} sharing a structured layout and storage zones fully covered by
shelves.
We divide the instance set into three groups, namely, \emph{small}, \emph{medium}, and \emph{large},
according to their increasing complexity.
To be more precise,
their layout sizes are
$11\times 6$,
$19\times 9$,
$46\times 15$
with
16,
60,
320 storage locations, respectively.%
\footnote{For the reader's convenience,
  we give the corresponding generator calls and layouts in Table~\ref{tab:eval:sizes}.}
Besides layout size, we categorize instances by number of robots:
\begin{itemize}
\item for  small instances: increments of 2, 5, 8 and 11 robots,
\item for medium instances: increments of 5, 10, 15 and 19 robots, and
\item for  large instances: increments of 12, 23, 35 and 46 robots.
\end{itemize}
We consider 30 instances per robot increment, and hence 120 per size and 360 in total.

Our experiments ran under Linux on an AMD Opteron 6278 with 32 cores and 256 GB,
using \clingo~5.3 and \clingcon~3.3 in their standard configurations.
Each run was limited to one thread on one physical core, 1800s, and 8 GB.
Notably, each instance is bound by its smallest pre-calculated makespan,
which makes the instances particularly hard to solve.
This makespan is calculated by the incremental encodings mentioned in Section~\ref{sec:encoding}
subject to a timeout of~8h.
Whenever the smallest makespan cannot be determined within this time limit,
we also consider the instance as unsolvable in the domain at hand.
Also,
assignments are pre-computed and the same assignment is used for all encodings in the same domain.
The computation of feasible assignments is more or less instantaneous;
their optimization was cut after 300s (but this only happens with large instances).

Our results are summarized in Table~\ref{tab:eval:results:summary}.
\begin{table}[h]
  \centering
  \renewcommand{\arraystretch}{1}
  \begin{tabular}{|c|c|c|*{3}{r}|}
    \cline{1-6}
    \emph{domain}              & \emph{makespan}          & \emph{encoding} & \emph{small} & \emph{medium} & \emph{large} \\
    \cline{1-6}
    \multirow{4}{*}{\Mdom}     & \multirow{4}{*}{6/10/25} & \clingo\        & \HN{0(0)}    & \HN{0(0)}     & \HN{73(4)}   \\
                               &                          & \clingosplit\   & \HN{0(0)}    & 16(1)         & 591(14)      \\
                               &                          & \clingcon\      & \HN{0(0)}    & 37(0)         & 1168(52)     \\
                               &                          & \clingod{dl}    & \HN{0(0)}    & 193(1)        & 1648(96)     \\
    \cline{1-6}
    \multirow{4}{*}{\MdomAsg}  & \multirow{4}{*}{6/10/25} & \clingo\        & \HN{0(0)}    & \HN{0(0)}     & \HN{41(2)}   \\
                               &                          & \clingosplit\   & \HN{0(0)}    & \HN{0(0)}     & 763(27)      \\
                               &                          & \clingcon\      & \HN{0(0)}    & 36(0)         & 1163(49)     \\
                               &                          & \clingod{dl}    & \HN{0(0)}    & 86(1)         & 1679(102)    \\
    \cline{1-6}
    \multirow{2}{*}{\CMdom}    & \multirow{2}{*}{20/-/-}  & \clingo\        & 805(40)      & 1800(120)     & 1800(120)    \\
                               &                          & \clingcon\      & \HN{695(30)} & 1800(120)     & 1800(120)    \\
    \cline{1-6}
    \multirow{2}{*}{\CMdomAsg} & \multirow{2}{*}{21/35/-} & \clingo\        & \HN{23(1)}   & \HN{370(5)}   & 1800(120)    \\
                               &                          & \clingcon\      & 38(2)        & 459(15)       & 1800(120)    \\
    \cline{1-6}
    \multirow{2}{*}{\BMdom}    & \multirow{2}{*}{26/-/-}  & \clingo\        & 970(53)      & 1800(120)     & 1800(120)    \\
                               &                          & \clingcon\      & \HN{807(37)} & 1800(120)     & 1800(120)    \\
    \cline{1-6}
    \multirow{2}{*}{\BMdomAsg} & \multirow{2}{*}{26/39/-} & \clingo\        & \HN{12(0)}   & \HN{566(19)}  & 1800(120)    \\
                               &                          & \clingcon\      & 29(0)        & 623(25)       & 1800(120)    \\
    \cline{1-6}
    \multirow{2}{*}{\AMdom}    & \multirow{2}{*}{26/-/-}  & \clingo\        & 984(55)      & 1800(120)     & 1800(120)    \\
                               &                          & \clingcon\      & \HN{856(41)} & 1800(120)     & 1800(120)    \\
    \cline{1-6}
    \multirow{2}{*}{\AMdomAsg} & \multirow{2}{*}{26/39/-} & \clingo\        & \HN{12(0)}   & \HN{577(18)}  & 1800(120)    \\
                               &                          & \clingcon\      & 49(1)        & 625(22)       & 1800(120)    \\
    \cline{1-6}
  \end{tabular}
  \caption{Summary of experimental results in average run time and number of timeouts.}
  \label{tab:eval:results:summary}
\end{table}%
%

The first column gives the respective \asprilo{} domain;
experiments with prior task assignments are indicated by adding subscript \textbf{a}.
The second column gives the average makespan of each size.
For instance,
for \MdomAsg{} the average makespans are 6, 10, and 25 for small, medium and large instances;
`-' tells us that the pre-computation of the makespan timed out for all instances in the domain.
The third column reflects the respective approach by giving the target system of the encoding.
As detailed in Section~\ref{sec:encoding},
\clingo{} stands for the encoding in Listing~\ref{enc:M} and~\ref{enc:M:O} (and its extensions),
\clingosplit{} for the ASP encoding splitting coordinates,
\clingod{dl} for the encoding mapping coordinates on two integer variables subject to difference constraints,
and analogously
\clingcon{} but using linear constraints.
Notably, \clingcon's constraint processing is extended in \Adom{} to product units.
The last three columns give average run time in seconds followed by the number of timeouts in parenthesis
obtained for task planning;
the best one in terms of timeouts (and secondarily time) is bold faced.
Timeouts account for 1800s;
run-times less than 5s are marked as 0.%
\footnote{For the reader's convenience, we give more fine-grained results in
  Table~\ref{tab:eval:res-small}, \ref{tab:eval:res-medium} and \ref{tab:eval:res-large}
  in the appendix.
  As mentioned, the full details of our experiments are given at \url{http://potassco.org/asprilo}.}

All encodings, except for the one related to \clingo{}, split positional coordinates.
While this is necessary with \clingcon{} and \clingod{dl} due to their use of integer variables,
\clingosplit{} allows us to examine the effect of splitting in plain ASP.
Contrary to our expectations,
this technique hardly led to performance improvements in our experiments.
Specifically,
for \Mdom\ and \MdomAsg{}, the \clingo{} encoding beats all others by a notable margin.
Although the other encodings produce smaller ground programs in these domains, in case of \clingcon{} and \clingod{dl} even by an order of magnitude,
the resulting number of constraints seems to increase.\footnote{This data was not available for \clingod{dl}.}
This changes for \AMdom, \BMdom, and \CMdom{},
where \clingcon{} takes the lead;
and roles switch again when
prior task assignments are used as in \AMdomAsg{}, \BMdomAsg, and \CMdomAsg.
We conjecture that this change is due to the reduction in problem size caused by the added task assignment (see below),
and that \clingcon's solving techniques gain more effect on larger instances.
However,
the catch-up effect of \clingcon{} in this context cannot clearly be traced back to coordinate splitting only.
Interestingly, the small integer domains disable \clingcon's lazy propagation,
so that all constraints are compiled out by using the order encoding~\cite{tatakiba09a}.
In fact, this offers a higher propagation strength than the direct encoding used in the \clingo{} encoding.

In the simplistic \Mdom{} domain,
all instance sizes are almost completely solvable.
Timeouts only occur for larger instances with 23 or more robots.
The transition to \CMdom{}
severely increases the average minimal makespan and therefore the run time.
In particular,
only small instances for robot increments 2, 5 and 8 are solvable in time for the most part,
whereas all medium and large instances time out.
This limitation carries over to \BMdom{} and \AMdom.
The more time-consuming delivery action in \BMdom{} (and \AMdom) further increases the average minimal makespan in comparison to \CMdom.
Conversely,
the addition of product quantities by \AMdom{} has no further impact on the average plan lengths.
Likewise,
there is a notable linear increase of run times between \CMdom{} and \BMdom{} whereas
the difference between \BMdom{} and \AMdom{} is relatively small.
This to be expected since only singleton product units are considered in \AMdom.%
\footnote{Increasing product units need intensive experimentation and are thus left to future research.}
So, units are taken into account but not extensively manipulated.

Our analysis has already made it apparent that a ``one-encoding-fits-all'' approach fails to scale on industrial size instances.
On the other hand,
the design of the instance generator has revealed how the compartmentalization of sources of combinatorics
allows for effectively reducing the overall difficulty of a problem.
This is confirmed by our experiments on the decoupling of task assignment and planning.
The provision of assignments improved scalability in each domain.
Especially for \CMdomAsg, \BMdomAsg, and \AMdomAsg,
the average run times decrease by at least an order of magnitude and almost all small and medium instances up to robot
increment 15 become solvable within the time limit.%
\footnote{We have no explanation why encoding~\clingosplit{} deteriorates when adding task assignments.}
Let us detail this by examining the path finding in instance \lstinline{x11_y6_n66_r8_s16_ps1_pr16_u16_o8_N001.lp} in domain \AMdom{}
with and without decoupling.
Although the added assignment slightly increases the incoming program size from
774584~rules and 65294~atoms to
818321~rules and 68284~atoms,
the internal representation obtained after preprocessing is much smaller,
namely,
1400191~constraints over 295835~atoms versus
 389874~constraints over 102316~atoms.
That is, here, the assignment makes the number of constraints shrink by an order of magnitude and halves the number of variables.
This reduction is less pronounced with \clingcon, where the number of rules/atoms and constraints/atoms drops from
143137/70127, 677131/95934
to
130064/59957, 292149/70666.
Likewise,
the difficulty of search in \clingo{} is decreased by several orders of magnitude
from
10807510~choices and 562464~conflicts to
  107732~choices and  28330~conflicts.
Finally,
our efforts to obtain economic plans pay off,
and the pre-determined task assignments lead to no increase of the smallest makespans,
except for \CMdom{} where it increases from 20 to 21.
Obviously, the makespans' stability is due to the optimized task assignments.
However, the problem reduction as such is obtained with any assignment.
Considering the large performance gain,
optimized assignments constitute an interesting trade-off for solution quality.
Moreover, this trade-off is easy to control since optimization can be done in an anytime manner.


%
\section{Related work}
\label{sec:related}\label{sec:discussion}

Our motivation is very similar to the early work on TheoryBase~\cite{chmamitr95a},
which provided our community with
a systematic way of creating logic programs from graph-based combinatorial problems.
Recently, for instance,
more application-oriented combinatorial problems 
were proposed in \cite{abgemuscwo15c,baiokascsotawa16a,aldoma17a},
dealing with shift planning, course timetabling, and nurse scheduling, respectively.
Unlike these static problems,
the decision support system for the space shuttle~\cite{nobagewaba01a} and
the Ricochet Robots board game~\cite{gejokaobsascsc13a}
were both put forward as dynamic benchmark problems,
also involving multiple, collaboratively controlled robots.
And many more can be found in the repositories of last years' ASP competitions.
What distinguishes robotic intra-logistics is its multidimensional nature
that necessitates the integration of a great many of aspects.

At the core of many path finding problems lies the search for a route for an agent from an initial to a final location.
The \emph{multi-agent path finding} (MAPF) problem asks for a collision-free route for each agent such that the total makespan is minimal.
MAPF is related to many real-world applications but
already computationally intractable \cite{surynek10a}.
While in MAPF each agent is assigned a unique destination,
its \emph{anonymous} variant requires no assignment of agents to destinations~\cite{yulav12a}.
The problem domains of \asprilo\ are obviously related to multi-agent path finding.
More specifically, the \asprilo{} domain \Mdom{} corresponds to anonymous MAPF.
Each order is uniquely associated with a destination shelf and there is no pre-assignment of a robot to an order.
Robots can freely reach any destination shelf.
Clearly, \Mdom{} is easily extended to cover non-anonymous MAPF by relating robots and orders.

\emph{Task assignment and path finding} (TAPF;~\citeNP{makoe16a}) is a generalization of MAPF.
TAPF groups agents into teams.
Although teams are (non-anonymously) pre-assigned to groups of destinations,
any robot in the team can be (anonymously) selected for a destination in the assigned group.
G-TAPF \cite{ngobsoscye17a} is a generalization of TAPF aiming at more realistic settings
by
allowing the number of tasks to be greater than the number of agents and
considering deadlines, orderings, and checkpoints.
That is,
deadlines are associated with order lines,
orders are completed in a pre-defined ordering and all lines in a single order need to be fulfilled before any line of another order is completed,
and
while fulfilling an order, a robot is required to go through a sequence of locations, called checkpoints.
Note that \asprilo's generic instance format can be easily extended to handle key concepts of TAPF and G-TAPF
(like robot teams, deadlines, or checkpoints) and
we plan to offer corresponding scenarios in the near future.
Regarding previous uses of ASP,
\cite{erkiozsc13a} address several aspects of multi-agent path finding problems.
The \asprilo{} framework is obviously beneficial for boosting such work.

\citeN{malikuko17a} address an online version of path finding,
where not all destination tasks%
\footnote{Actually, this work also uses pick-up and delivery tasks to simulate a warehouse system.}
are given initially but may arrive over time.
In view of the availability of multi-shot ASP solving,
the treatment of online versions of \asprilo's problem domains constitutes an important future line of research for us.
Several emerging requirements of online intra-logistic problems have already been taken into account in the design of \asprilo.

\section{Conclusion}
\label{sec:conclusion}

The current \asprilo{} system provides a basic infrastructure for experimental studies of dynamic systems.
As is, it already presents a number of challenges to ASP-based approaches for solving intra-logistics problems,
whenever scalability and efficiency are at stake.
Abstraction and distribution appear to be promising avenues of future research towards scalability.
Thanks to its open design, it can be extended in various ways.
Dealing with preferences, different types of orders, and uncertainty provide additional significant challenges
that might require new insights or approaches.
Last but not least,
online approaches to solving intra-logistics problems are interesting and might even be necessary.

To sum up,
we have presented \asprilo, a framework for experimental studies of dynamic systems, specifically in the intra-logistics domains.
We discussed our design decisions and detailed the different components of \asprilo:
its benchmark generator, solution checker, visualizer, and sample encodings.
We illustrated \asprilo's utility of providing an experimental platform for solving various problems in robotic intra-logistics
and revealed significant challenges to our community in closing the gap to industrial-scale applications.

Videos illustrating \asprilo{} in domains \Adom{} and \Mdom{} can be found \url{https://youtu.be/ifYKHIvdnjw} and
\url{https://youtu.be/GHRwpWzL0j8}, respectively.
%
%

\paragraph{Acknowledgments.}
This work was partially supported by
The Scientific and Technological Research Council of Turkey (TUBITAK) under project 117C044
and
DFG grant SCHA 550/9.


\bibliographystyle{acmtrans}


\appendix
\newpage
\section{Benchmark instance layout}
\label{sec:eval:sizes}
    \begin{table}[h]
      \centering \footnotesize
      \renewcommand{\arraystretch}{0.1}
      \lstset{basicstyle=\fontsize{6}{8}\ttfamily,
        numbers=none,frame=single,linewidth=10.2cm}
      \begin{tabular}{cc}
        Name                            & Generator Call and Resulting Layout                              \\
        \hline\hline
        \multirow{2}{*}{\emph{small:}}  &
\hspace{-1.4cm} 
\begin{lstlisting}
gen -x 11 -y  6 -X  4 -Y  2 -p  1 -s  16 -P  16 -u  16 -H --prs 1 *@\textcolor{blue}{-r 2 -o 2}@*
\end{lstlisting}
                                                                                                           \\[0.2cm]
                                        & \includegraphics[width=0.15\textwidth]{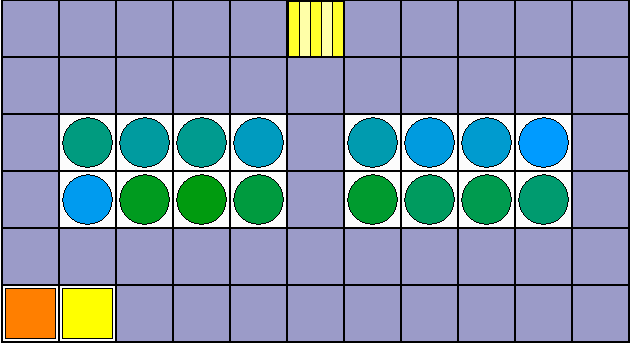} \\
        \hline
        \multirow{2}{*}{\emph{medium:}} &
\hspace{-1.4cm}
\begin{lstlisting}
gen -x 19 -y  9 -X  5 -Y  2 -p  3 -s  60 -P  60 -u  60 -H --prs 1 *@\textcolor{blue}{-r 5 -o 5}@*
\end{lstlisting}
                                                                                                           \\[0.2cm]
                                        & \includegraphics[width=0.3\textwidth]{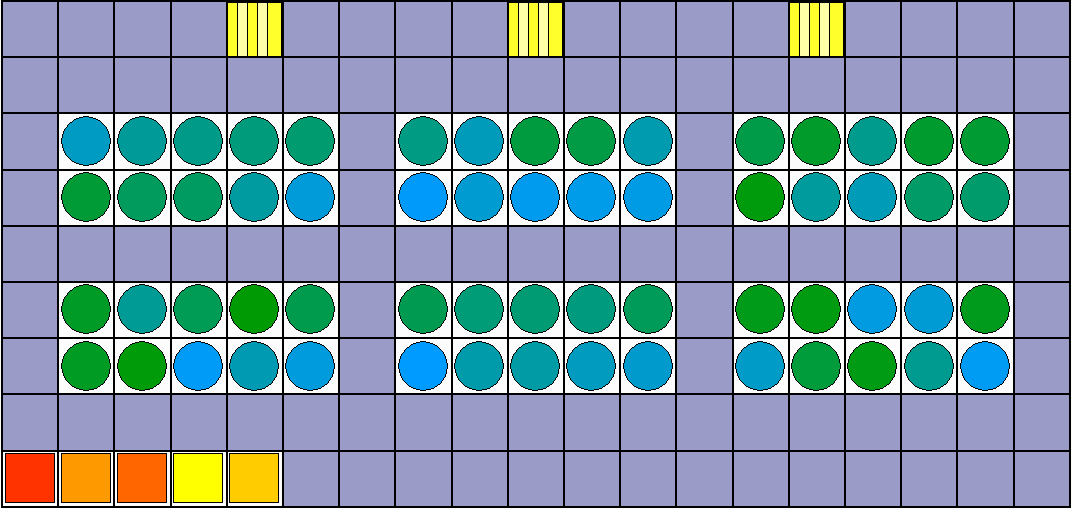} \\
        \hline
        \multirow{2}{*}{\emph{large:}}  &
\hspace{-1.4cm}
\begin{lstlisting}
gen -x 46 -y 15 -X  8 -Y  2 -p 10 -s 320 -P 320 -u 320 -H --prs 1 *@\textcolor{blue}{-r 12 -o 12}@*
\end{lstlisting}
                                                                                                           \\[0.2cm]
                                        & \includegraphics[width=0.8\textwidth]{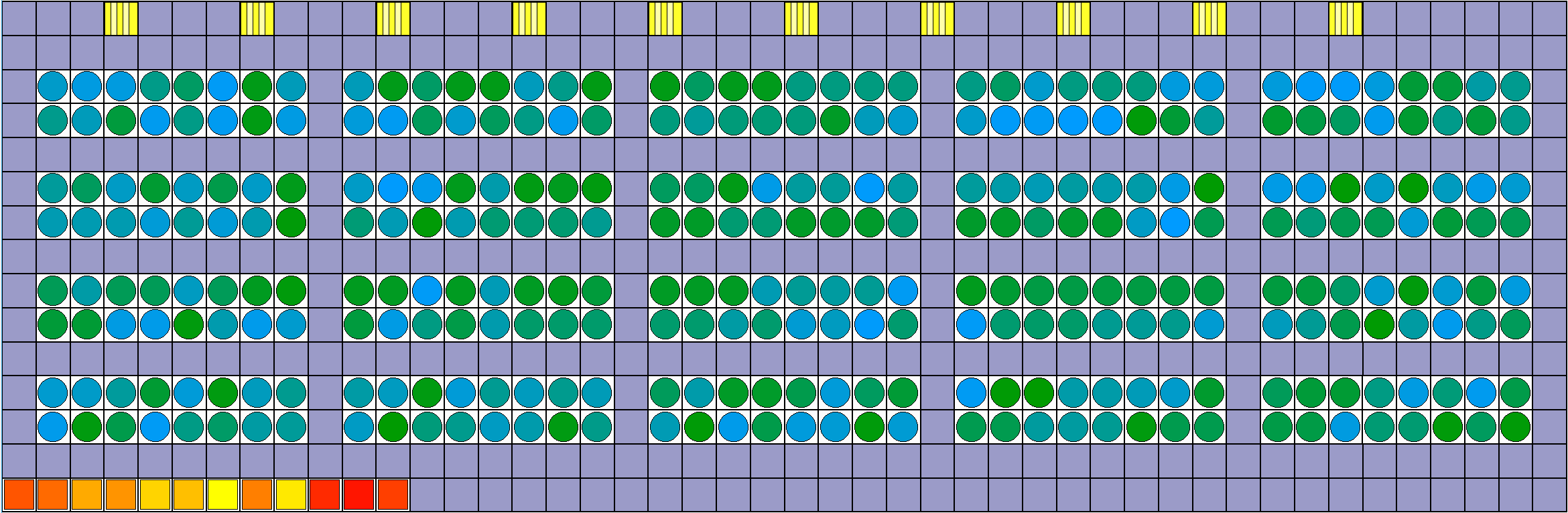}  \\
        \hline\hline
      \end{tabular}
      \caption{Instance layout sizes specified by generator calls and resulting layouts;
        particularly, the depicted calls show the first robot increment per layout size with the
        related parameters for robots and orders colored in blue.}
      \label{tab:eval:sizes}
    \end{table}

\newpage
\section{Detailed evaluation results}
\label{app:evaluation}

  \begin{table}[h]
    \centering \footnotesize
    \renewcommand{\arraystretch}{1}
    \begin{tabular}{|c|c|c| *{4}{r}|}
      \cline{1-7}
      \emph{domain}              & \emph{makespan}     & \emph{encoding} & \emph{2 robots} & \emph{5 robots} & \emph{8 robots} & \emph{11 robots} \\
      \cline{1-7}
      \multirow{4}{*}{\Mdom}     & \multirow{4}{*}{6}  & \clingo\        & \HN{0(0)}       & \HN{0(0)}       & \HN{0(0)}       & \HN{0(0)}        \\
                                 &                     & \clingosplit\   & \HN{0(0)}       & \HN{0(0)}       & \HN{0(0)}       & \HN{0(0)}        \\
                                 &                     & \clingcon\      & \HN{0(0)}       & \HN{0(0)}       & \HN{0(0)}       & \HN{0(0)}        \\
                                 &                     & \clingod{dl}\   & \HN{0(0)}       & \HN{0(0)}       & \HN{0(0)}       & \HN{0(0)}        \\
      \cline{1-7}
      \multirow{4}{*}{\MdomAsg}  & \multirow{4}{*}{6}  & \clingo\        & \HN{0(0)}       & \HN{0(0)}       & \HN{0(0)}       & \HN{0(0)}        \\
                                 &                     & \clingosplit\   & \HN{0(0)}       & \HN{0(0)}       & \HN{0(0)}       & \HN{0(0)}        \\
                                 &                     & \clingcon\      & \HN{0(0)}       & \HN{0(0)}       & \HN{0(0)}       & \HN{0(0)}        \\
                                 &                     & \clingod{dl}\   & \HN{0(0)}       & \HN{0(0)}       & \HN{0(0)}       & \HN{0(0)}        \\
      \cline{1-7}
      \multirow{2}{*}{\CMdom}    & \multirow{2}{*}{20} & \clingo\        & 46(0)           & 751(9)          & 623(1)          & 1800(30)         \\
                                 &                     & \clingcon\      & \HN{17(0)}      & \HN{580(5)}     & \HN{411(0)}     & \HN{1772(25)}    \\
      \cline{1-7}
      \multirow{2}{*}{\CMdomAsg} & \multirow{2}{*}{21} & \clingo\        & \HN{0(0)}       & 4(0)            & 67(1)           & \HN{21(0)}       \\
                                 &                     & \clingcon\      & 61(1)           & 4(0)            & \HN{9(0)}       & 80(1)            \\
      \cline{1-7}
      \multirow{2}{*}{\BMdom}    & \multirow{2}{*}{26} & \clingo\        & 37(0)           & 398(1)          & 1647(22)        & 1800(30)         \\
                                 &                     & \clingcon\      & \HN{15(0)}      & \HN{254(0)}     & \HN{1159(7)}    & 1800(30)         \\
      \cline{1-7}
      \multirow{2}{*}{\BMdomAsg} & \multirow{2}{*}{26} & \clingo\        & \HN{0(0)}       & \HN{5(0)}       & 23(0)           & \HN{22(0)}       \\
                                 &                     & \clingcon\      & \HN{0(0)}       & 6(0)            & \HN{15(0)}      & 94(0)            \\
      \cline{1-7}
      \multirow{2}{*}{\AMdom}    & \multirow{2}{*}{26} & \clingo\        & 52(0)           & 422(2)          & 1662(23)        & 1800(30)         \\
                                 &                     & \clingcon\      & \HN{18(0)}      & \HN{303(1)}     & \HN{1304(10)}   & 1800(30)         \\
      \cline{1-7}
      \multirow{2}{*}{\AMdomAsg} & \multirow{2}{*}{26} & \clingo\        & \HN{0(0)}       & \HN{0(0)}       & 25(0)           & \HN{20(0)}       \\
                                 &                     & \clingcon\      & \HN{0(0)}       & \HN{0(0)}       & \HN{16(0)}      & 175(1)           \\
      \cline{1-7}
    \end{tabular}
    \caption{Average run time and number of time outs for small instances.}
    \label{tab:eval:res-small}
  \end{table}

  \begin{table}[h]
    \centering \footnotesize
    \renewcommand{\arraystretch}{1}
    \begin{tabular}{|c|c|c| *{4}{r}|}
      \cline{1-7}
      \emph{domain}              & \emph{makespan}     & \emph{encoding} & \emph{5 robots} & \emph{10 robots} & \emph{15 robots} & \emph{19 robots} \\
      \cline{1-7}
      \multirow{4}{*}{\Mdom}     & \multirow{4}{*}{10} & \clingo\        & \HN{0(0)}       & \HN{0(0)}        & \HN{0(0)}        & \HN{0(0)}        \\
                                 &                     & \clingosplit\   & \HN{0(0)}       & \HN{0(0)}        & 61(1)            & 3(0)             \\
                                 &                     & \clingcon\      & \HN{0(0)}       & \HN{0(0)}        & 11(0)            & 135(0)           \\
                                 &                     & \clingod{dl}\   & 5(0)            & 120(1)           & 180(0)           & 467(0)           \\
      \cline{1-7}
      \multirow{4}{*}{\MdomAsg}  & \multirow{4}{*}{10} & \clingo\        & \HN{0(0)}       & \HN{0(0)}        & \HN{0(0)}        & \HN{0(0)}        \\
                                 &                     & \clingosplit\   & \HN{0(0)}       & \HN{0(0)}        & \HN{0(0)}        & \HN{0(0)}        \\
                                 &                     & \clingcon\      & \HN{0(0)}       & \HN{0(0)}        & 16(0)            & 128(0)           \\
                                 &                     & \clingod{dl}\   & 18(0)           & 150(1)           & 76(0)            & 103(0)           \\
      \cline{1-7}
      \multirow{2}{*}{\CMdom}    & \multirow{2}{*}{-}  & \clingo\        & 1800(30)        & 1800(30)         & 1800(30)         & 1800(30)         \\
                                 &                     & \clingcon\      & 1800(30)        & 1800(30)         & 1800(30)         & 1800(30)         \\
      \cline{1-7}
      \multirow{2}{*}{\CMdomAsg} & \multirow{2}{*}{35} & \clingo\        & 50(0)           & \HN{106(0)}      & \HN{310(0)}      & \HN{1015(5)}     \\
                                 &                     & \clingcon\      & \HN{41(0)}      & 116(0)           & 512(2)           & 1169(13)         \\
      \cline{1-7}
      \multirow{2}{*}{\BMdom}    & \multirow{2}{*}{-}  & \clingo\        & 1800(30)        & 1800(30)         & 1800(30)         & 1800(30)         \\
                                 &                     & \clingcon\      & 1800(30)        & 1800(30)         & 1800(30)         & 1800(30)         \\
      \cline{1-7}
      \multirow{2}{*}{\BMdomAsg} & \multirow{2}{*}{39} & \clingo\        & 53(0)           & \HN{151(0)}      & \HN{569(1)}      & \HN{1492(18)}    \\
                                 &                     & \clingcon\      & \HN{47(0)}      & 159(0)           & 709(3)           & 1579(22)         \\
      \cline{1-7}
      \multirow{2}{*}{\AMdom}    & \multirow{2}{*}{-}  & \clingo\        & 1800(30)        & 1800(30)         & 1800(30)         & 1800(30)         \\
                                 &                     & \clingcon\      & 1800(30)        & 1800(30)         & 1800(30)         & 1800(30)         \\
      \cline{1-7}
      \multirow{2}{*}{\AMdomAsg} & \multirow{2}{*}{39} & \clingo\        & 59(0)           & 159(0)           & 594(1)           & \HN{1497(17)}    \\
                                 &                     & \clingcon\      & \HN{48(0)}      & \HN{143(0)}      & \HN{701(0)}      & 1608(22)         \\
      \cline{1-7}
    \end{tabular}
    \caption{Average run time and number of timeouts for medium-sized instances.}
    \label{tab:eval:res-medium}
  \end{table}

  \begin{table}[h]
    \centering \footnotesize
    \renewcommand{\arraystretch}{1}
    \begin{tabular}{|c|c|c| *{4}{r}|}
      \cline{1-7}
      \emph{domain}              & \emph{makespan}     & \emph{encoding} & \emph{12 robots} & \emph{23 robots} & \emph{35 robots} & \emph{46 robots} \\
      \cline{1-7}
      \multirow{4}{*}{\Mdom}     & \multirow{4}{*}{25} & \clingo\        & \HN{15(0)}       & \HN{79(1)}       & \HN{15(0)}       & \HN{186(3)}      \\
                                 &                     & \clingosplit\   & 205(0)           & 1007(7)          & 808(4)           & 347(3)           \\
                                 &                     & \clingcon\      & 119(0)           & 1108(3)          & 1748(27)         & 1700(22)         \\
                                 &                     & \clingod{dl}\   & 1193(6)          & 1800(30)         & 1800(30)         & 1800(30)         \\
      \cline{1-7}
      \multirow{4}{*}{\MdomAsg}  & \multirow{4}{*}{25} & \clingo\        & \HN{13(0)}       & \HN{18(0)}       & \HN{10(0)}       & \HN{125(2)}      \\
                                 &                     & \clingosplit\   & 249(1)           & 1282(15)         & 1096(7)          & 427(4)           \\
                                 &                     & \clingcon\      & 162(1)           & 1053(0)          & 1734(25)         & 1706(23)         \\
                                 &                     & \clingod{dl}\   & 1316(12)         & 1800(30)         & 1800(30)         & 1800(30)         \\
      \cline{1-7}
      \multirow{2}{*}{\CMdom}    & \multirow{2}{*}{-}  & \clingo\        & 1800(30)         & 1800(30)         & 1800(30)         & 1800(30)         \\
                                 &                     & \clingcon\      & 1800(30)         & 1800(30)         & 1800(30)         & 1800(30)         \\
      \cline{1-7}
      \multirow{2}{*}{\CMdomAsg} & \multirow{2}{*}{-}  & \clingo\        & 1800(30)         & 1800(30)         & 1800(30)         & 1800(30)         \\
                                 &                     & \clingcon\      & 1800(30)         & 1800(30)         & 1800(30)         & 1800(30)         \\
      \cline{1-7}
      \multirow{2}{*}{\BMdom}    & \multirow{2}{*}{-}  & \clingo\        & 1800(30)         & 1800(30)         & 1800(30)         & 1800(30)         \\
                                 &                     & \clingcon\      & 1800(30)         & 1800(30)         & 1800(30)         & 1800(30)         \\
      \cline{1-7}
      \multirow{2}{*}{\BMdomAsg} & \multirow{2}{*}{-}  & \clingo\        & 1800(30)         & 1800(30)         & 1800(30)         & 1800(30)         \\
                                 &                     & \clingcon\      & 1800(30)         & 1800(30)         & 1800(30)         & 1800(30)         \\
      \cline{1-7}
      \multirow{2}{*}{\AMdom}    & \multirow{2}{*}{-}  & \clingo\        & 1800(30)         & 1800(30)         & 1800(30)         & 1800(30)         \\
                                 &                     & \clingcon\      & 1800(30)         & 1800(30)         & 1800(30)         & 1800(30)         \\
      \cline{1-7}
      \multirow{2}{*}{\AMdomAsg} & \multirow{2}{*}{-}  & \clingo\        & 1800(30)         & 1800(30)         & 1800(30)         & 1800(30)         \\
                                 &                     & \clingcon\      & 1800(30)         & 1800(30)         & 1800(30)         & 1800(30)         \\
      \cline{1-7}
    \end{tabular}
    \caption{Average run time and number of timeouts for large instances.}
    \label{tab:eval:res-large}
  \end{table}


\end{document}